\documentclass{article}

\usepackage{url}
\usepackage{booktabs}
\usepackage{XCharter}
\usepackage{graphicx}
\usepackage{multirow}
\usepackage{multicol}
\usepackage{makecell}
\usepackage{amsmath}
\usepackage{amsfonts}
\usepackage[bb=px]{mathalpha}
\usepackage[]{algorithm}
\usepackage[noend]{algpseudocode}
\usepackage[table]{xcolor}
\usepackage{todonotes}
\usepackage{relsize}
\usepackage{setspace}
\usepackage[numbers]{natbib}

\usepackage{tikz}
\usetikzlibrary{shapes.geometric, arrows, trees, arrows.meta, positioning}
\usepackage{stackengine}

\usepackage[bb=px]{mathalpha}
\usepackage{multirow}
\usepackage{multicol}
\usepackage{makecell}
\usepackage{subcaption}
\newcommand{\citationneeded}[1][]{\textsuperscript{\color{blue} [citation needed]}}

\newtheorem{definition}{Definition}

\newtheorem{assumption}{Assumption}

\newcommand{\probf}{\ensuremath{P}}
\newcommand{\rewardf}{\ensuremath{R}}

\newcommand{\budget}{\ensuremath{b}}
\newcommand{\goals}{\ensuremath{G}}

\newcommand{\judgef}{\ensuremath{J} }
\newcommand{\estimateType}{\ensuremath{\mathcal{W}}}

\newcommand{\policy}{\ensuremath{\pi} }
\newcommand{\gatherBasef}{\ensuremath{Q}}
\newcommand{\gatherf}{\ensuremath{\mathcal{Q}}}
\newcommand{\estGEQ}{\ensuremath{\succeq} }

\newcommand{\moralTheory}{\ensuremath{\langle \estimateType, \judgef, \gatherBasef, \preceq, \approx \rangle}}
\newcommand{\estf}{\ensuremath{W}}

\newcommand{\vgatherf}{\boldsymbol{\gatherf}}

\newcommand{\CQ}{\ensuremath{\Psi}}
\newcommand{\heuristic}{\ensuremath{\eta}}

\newcommand{\attack}{\rotatebox[origin=c]{90}{$\dagger$}}

\newcommand{\numbermode}[1]{\ensuremath{#1}}

\newcommand{\ProbSpAqSp}{\numbermode{0.4}}
\newcommand{\ProbSpAqSq}{\numbermode{0.6}}

\newcommand{\ProbSpAwSw}{\numbermode{0.6}}
\newcommand{\ProbSpAwSe}{\numbermode{0.15}}
\newcommand{\ProbSpAwSr}{\numbermode{0.15}}
\newcommand{\ProbSpAwSt}{\numbermode{0.1}}

\newif\ifcomments
\newif\ifshowremoved

\usepackage{xcolor}
\usepackage[absolute]{textpos}
\usepackage{everypage}

\definecolor{lightblue}{RGB}{210,210,225}
\definecolor{lightred}{RGB}{225,210,210}
\definecolor{lightgreen}{RGB}{210,225,210}
\definecolor{lightyellow}{RGB}{225,222,200}
\definecolor{lightpurple}{RGB}{225,210,225}
\definecolor{warningyellow}{RGB}{247, 245, 187}
\definecolor{darkergreen}{RGB}{0,64,0}
\definecolor{darkred}{RGB}{128,0,0}
\definecolor{darkblue}{RGB}{0,0,139}
\definecolor{darkgreen}{RGB}{0,128,0}
\definecolor{darkpurple}{RGB}{128,0,128}
\definecolor{warningorange}{RGB}{124, 81, 0}
\definecolor{eyecancerpink}{rgb}{1.0, 0.0, 1.0}
\definecolor{radiationyellow}{rgb}{0.8, 1.0, 0.0}

\makeatletter
\def\THICKhrulefill{\leavevmode \leaders \hrule height 5pt\hfill \kern \z@}
\makeatother

\usepackage{balance} 

\title{\vspace{-5em}\textbf{Uncertain Machine Ethics Planning}}

\author{
  \scalebox{0.8}{
  \makebox[\textwidth][c]{%
    \begin{tabular}{cc}
      \begin{tabular}{c}
        Simon Kolker \\
        University of Manchester, UK \\
        \texttt{simon.kolker@manchester.ac.uk}
      \end{tabular}
      &
      \begin{tabular}{c}
        Louise A. Dennis \\
        University of Manchester, UK \\
        \texttt{louise.dennis@manchester.ac.uk}
      \end{tabular}
      \\
      \\
      \begin{tabular}{c}
        Ramon Fraga Pereira \\
        University of Manchester, UK \\
        \texttt{ramon.fragapereira@manchester.ac.uk}
      \end{tabular}
      &
      \begin{tabular}{c}
        Mengwei Xu \\
        Newcastle University, UK \\
        \texttt{mengwei.xu@newcastle.ac.uk}
      \end{tabular}
    \end{tabular}
  }
  }
}

\date{}

\begin{document}

\maketitle 

\let\thefootnote\relax
\footnotetext{The original version of this paper has been accepted by the 24th International Conference on Autonomous Agents and Multiagent Systems (AAMAS 2025) as a Full Paper. It is available by open access under the terms of the Creative Commons Attribution Licence (CC BY 4.0) \url{https://creativecommons.org/licenses/by/4.0}}

\begin{abstract}
  Machine Ethics decisions should consider the implications of uncertainty over decisions.
  Decisions should be made over sequences of actions to reach preferable outcomes long term.
  The evaluation of outcomes, however, may invoke one or more moral theories, which might have conflicting judgements.
  Each theory will require differing representations of the ethical situation.
  For example, \textit{Utilitarianism} measures numerical values, \textit{Deontology} analyses duties, and \textit{Virtue Ethics} emphasises moral character.
  While balancing potentially conflicting moral considerations, decisions may need to be made, for example, to achieve morally neutral goals with minimal costs.
  In this paper, we formalise the problem as a \textit{Multi-Moral Markov Decision Process} and a \textit{Multi-Moral Stochastic Shortest Path Problem}.
  We develop a heuristic algorithm based on \textit{Multi-Objective AO*}, utilising Sven-Ove Hansson's \textit{Hypothetical Retrospection} procedure for ethical reasoning under uncertainty.
  Our approach is validated by a case study from Machine Ethics literature: the problem of whether to steal insulin for someone who needs it.
  \end{abstract}



\section{Introduction}
As our lives are increasingly impacted by the decisions of machines \cite{Robophilosophy_Wallach14}, one objective of \textit{Machine Ethics} is to develop computational techniques to incorporate ethical behaviour with decision-making, preserving human values into the future~\cite{MachineEthics_AllenWS06}.
A major concern for Machine Ethics comes from situations with uncertainty over outcomes, commonplace in the real world.
Most existing approaches either assume actions are deterministic or they use expected utility maximisation~\cite{tolmeijer2020implementations}.
This can yield disagreeable results (see Section 3 and \cite{brundage2014limitations}), demanding a more sophisticated handling of \emph{outcome uncertainty}. 
Decisions are further complicated by \textit{moral uncertainty}~\cite{ecoffet2021reinforcement}. Philosophy has no universal consensus over which decisions are correct in all situations: moral theories may conflict, leading to \emph{moral dilemmas}.
While we defer the choice of theories to \emph{stakeholders} who must approve a system before deployment (owners, regulators, or affected individuals), there are likely to be conflicts or uncertainty between stakeholder approved theories. Implementations must handle these situations gracefully.

Another issue is the integration of ethical behaviour into existing systems with non-moral goals.
For example, an autonomous vehicle must balance risk to other road users with reaching its destination.

As the \textit{Automated Planning} community demonstrates, many environments require sequences of actions (action plans or policies) to achieve desired goals and the most rewarding outcomes~\cite{2013GeffnerBonetBook}. Machine Ethics implementations, therefore, should consider future decisions, rather than selecting immediately preferred actions greedily as is the case in most existing systems~\cite{tolmeijer2020implementations}. 

\begin{figure}[t]
    \centering
    \scalebox{0.8}{
        \usetikzlibrary{positioning, arrows.meta}
\begin{tikzpicture}[
  grow=down,
  level 1/.style={sibling distance=4cm, level distance=1cm},
  level 2/.style={sibling distance=2cm, level distance=1.5cm},
  level 3/.style={sibling distance=2cm, level distance=2cm},
  edge from parent/.style={draw,-latex},
  root/.style={circle,draw, minimum size=1cm, align=center},
  outcome/.style={circle,draw, minimum size=1cm, align=center, text width=1cm, font=\footnotesize}, 
  parallelogram/.style={draw, trapezium, trapezium angle=70, minimum height=1cm,  trapezium stretches=true, trapezium stretches body=true, align=center, text width=1cm, font=\footnotesize}, 
  attack/.style={line width = 1pt, -{Latex[width=3mm]}},
  argument/.style={draw, rectangle, align=center, text width=3.2cm}
]

\node[draw, rectangle, align=center] at (0,0) (start) {CHOOSE POLICY};

\node[draw, circle, text width=0.4cm, align=center, below left=-0.2cm and 1.25cm of start] (pi_1) {$\policy$};

\node[draw, circle, text width=0.4cm, align=center, below right=-0.2cm and 1.25cm of start](pi_2) {$\policy'$};

\node[argument, below=0.5cm of pi_1] (arg1) {From $s_0$, $\policy$ was correct, given greater beauty care production with probability 1.};

\node[argument, below left=0.555cm and -0.2cm of pi_2] (arg2) {From $s_0$, $\policy'$ was correct, resulting in a new drug with probability $0.5$.};

\node[argument, below right=0.52cm and -0.2cm of pi_2] (arg3) {From $\boldsymbol{s_0}$, $\policy'$ was correct, resulting in no new drug with probability $0.5$.};

\draw[->, line width=0.3mm, -{Triangle[width=3mm]}] (start.west) to (pi_1.east);
\draw[->, line width=0.3mm, -{Triangle[width=3mm]}] (start.east) to (pi_2.west);

\draw[->, line width=0.3mm, -{Triangle[width=2mm]}] (pi_1.south) .. controls +(0,-0.5) and +(0,0.5) .. (arg1.north);
\draw[->, line width=0.3mm, -{Triangle[width=2mm]}] (pi_2.south) .. controls +(0,-0.5) and +(0,0.5) .. (arg2.north);
\draw[->, line width=0.3mm, -{Triangle[width=2mm]}] (pi_2.south) .. controls +(0,-0.5) and +(0,0.5) .. (arg3.north);

\draw[<-, line width=0.5mm, color=red, -{Stealth[width=3mm]}] 
    (arg2.south) .. controls +(-0.5,-1.5) and +(0.8,-1.3) .. (arg1.south) 
    node[pos=0.53, sloped, font=\smaller, align=center, yshift=-0.05cm] 
    {\begin{tabular}{c} NEGATIVE \\[0.3em] RETROSPECTION \end{tabular}};

\end{tikzpicture}
    }
    \caption{A company may produce a beauty care product or research a drug with 50\% probability of saving lives. Beauty care has negative retrospection (like regret) for missing the chance to save lives; if research is unsuccessful, there is no negative retrospection since risk was accepted at decision-time. Arguments are generated from each outcome. Directed edges are attacks indicating negative retrospection.}
    \label{fig:MEHR example}
\end{figure}
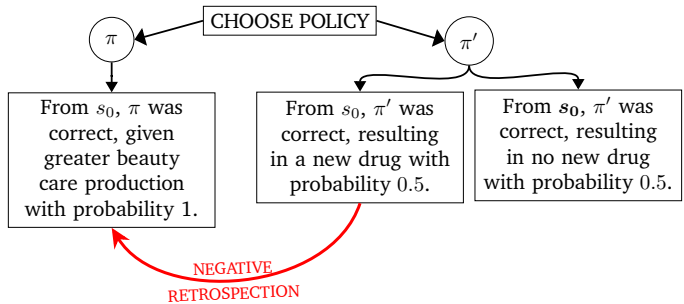
\textit{Hypothetical Retrospection} is an ethical procedure for handing outcome uncertainty~\cite{hansson2013ethics}. It suggests we judge decisions by the circumstances under which they were made.
In \cite{kolker2023uncertain}, \textit{Machine Ethics Hypothetical Retrospection} (MEHR) is formalised as a framework for single action decisions with outcome uncertainty and multiple moral theories under  moral uncertainty. Arguments are generated in support of each action from the hypothetical perspective of their outcomes. An argument is attacked if there is negative retrospection for missing an ethically preferred outcome. An action that minimises non-acceptability across arguments is selected. There is an example in Figure~\ref{fig:MEHR example}, details in Section~\ref{sec:background}.

In this paper, we formalise uncertain ethical planning problems as a variant of the \emph{Markov Decision Process} (MDP).
Our \emph{Multi-Moral Markov Decision Process} (MMMDP) may contain a number of potentially conflicting moral theories.
Stakeholders may express preference or moral uncertainty over theories with a \emph{weak lexicographic ordering}.
A theory's space of morally relevant information is general and inclusive to many ethical perspectives.
We extend our formalism to the \emph{Multi-Moral Stochastic Shortest Path Problem} (MMSSP) for non-moral situations. Preferable solutions act ethically subject to goals and costs.
We produce an offline heuristic planning algorithm to solve these problems based on AO*~\cite{NILSSON198099}, Multi-Objective heuristic planning~\cite{chen2023heuristic} and MEHR~\cite{kolker2023uncertain}.
We evaluate the algorithm with a case study from Ethics and Machine Ethics literature: the problem of whether to steal insulin for someone who needs it~\cite{atkinson2008addressing}.
\section{Background}
\label{sec:background}
\subsubsection*{Markov Decision Process}
Uncertain decision making problems can be described as a \textit{Finite Horizon Markov Decision Process} (MDP). 
We represent them as a tuple $\langle S, A, \probf, R, H \rangle$, where
$S$ is a finite set of states, and $A$ is a finite set of actions.
The state space is traversed through discrete sequential transitions.
The probabilistic transition function $\probf : S \times A \times S \rightarrow [0,1]$ defines the probability one state transitions to another via an action.
Transitions are valued by a reward function $\rewardf: S \times A \times S \rightarrow \mathbb{R}$.
An optimal solution is a non-stationary policy $\policy: S \times \{0,...,H-1\} \rightarrow A$ that maps a state and timestamp to an action with maximum expected reward. 
The horizon $H\in \mathbb{N}$ is the number of state transitions before the problem terminates.
Policies are evaluated with a non-stationary value function defined recursively: $V^\policy(s,t)=\sum_{s'\in S} \probf(s, \policy(s,t), s')[R(s,\policy(s,t),s') + V^\policy(s',t+1)]$.
By \citeauthor{Bellman_DynamicProgramming_57}'s optimality principle~\cite{Bellman_DynamicProgramming_57},if a policy's value can be measured by its expected additive utility, then there exists a policy $\pi^*$ optimal at all sates and it satisfies the \textit{\citeauthor{Bellman_DynamicProgramming_57} optimality equation} $V^{\policy^*}(s,t)=\max_{a\in A} \sum_{s' \in S} \probf(s,a,s')[R(s,a,s') + V^{\policy^*}(s',t+1)]$.
\textit{Stochastic Shortest Path} (SSP) problems are MDPs with goal states $G \subseteq S$ that are terminal, meaning there are no transitions away from them: $\forall_{s\in G, a \in A}, P(s,a,s)=1 \land R(s,a,s)=0$;
all other rewards are strictly negative. SSPs also have an initial state $s_0$.
Optimal policies are \emph{proper} if there is $1.0$ probability of reaching a goal from $s_0$. Improper policies have an expected cost of $\infty$.

\subsubsection*{MDP Algorithms}
MDPs can be solved using \textit{Dynamic Programming} algorithms, e.g. \textit{Policy Iteration} \cite{Howard_MDP_1960,Bertsekas_VI_PI_1995} or \textit{Value Iteration} (VI)~\cite{Bertsekas_VI_PI_1995}.
VI refines an arbitrary value function $V_0$ with
\emph{\citeauthor{Bellman_DynamicProgramming_57} backups}: iterative applications of the \citeauthor{Bellman_DynamicProgramming_57} optimality equation~\cite{Bellman_DynamicProgramming_57}.
Values propagate until the maximum absolute difference between values in $V_{i}$ and $V_{i+1}$ (the \textit{residual}) is below a constant $\epsilon > 0$ ($\epsilon$-consistency). 
This requires \textit{full sweeps} across the state space. 
\textit{Heuristic} algorithms (RTDP~\cite{AIJ_RTDP_BartoBS95}, AO*~\cite{NILSSON198099}, LAO*~\cite{AIJ_HansenZ01a_LAOStar} and LRTDP~\cite{LRTDP_BonetG03}) improve performance when there is a initial state $s_0$ by only iterating on reachable, high value states.
They also use a heuristic function to initialise $V_0$ to an estimate of each state's optimal reward.

\subsubsection*{Multi-Objective MDP}
Many decision scenarios cannot be modelled by a single reward function. A \textit{Multi-Objective Markov Decision Process} (MOMDP) generalises MDP rewards into vectors $\boldsymbol{R} : S \times A \times S \rightarrow \mathbb{R}^n$ for a number of objectives $n>0$.
A given policy may be optimal with respect to one objective, but not the others. Thus, we are interested in the set of \emph{undominated} policies: policies where no objective can be improved with no detriment to another objective.
For notation, we treat value functions interchangeably with ordered sets: $v_i=V(s_i)$ for $s_i\in S$.
Thus, a single-objective function $V$ dominates $U$ if $\forall{i\in [1,\ldots, n]}, v_i < u_i$, written $V \prec U$.
A set of vectors $\boldsymbol{V}$ dominates $\boldsymbol{U}$ if $\forall{V \in \boldsymbol{V}}$ there exists $U \in \boldsymbol{U}$ such that $U\prec V$.
Finally, a policy $\policy$ dominates $\policy'$ if $V^{\policy'} \prec V^\policy$. The set of undominated policies is the maximal set $\Pi^*$ where $\forall{\pi\in \Pi^*, \pi' \not\in \Pi^*} : V^{\pi} \not\prec V^{\pi'}$. 

MDP algorithms are adapted for MOMDPs using two distinct approaches~\cite{roijers2017multi}.
Outer-loop algorithms repeatedly run single-objective solvers on scalarised versions of the problem. Inner-loop approaches generalise the backup operation for multiple objectives e.g., the value function generalises neatly with vector addition $\boldsymbol{V}^\policy(s,t)=\sum_{s'\in S} \probf(s, \policy(s,t), s')[\boldsymbol{R}(s,\policy(s,t),s') \oplus \boldsymbol{V}^\policy(s',t+1)]$.
This paper adapts an inner-loop approach for heuristic planning~\cite{chen2023heuristic}.

\subsubsection*{Hypothetical Retrospection}
Hansson's \textit{Hypothetical Retrospection} procedure systematises human ethical decision-making in scenarios with uncertain outcomes~\cite{hansson2013ethics}.
\citeauthor{hansson2013ethics} suggests we extend our perspective with future perceptions of actions. He argues to view the ethics of decisions from perspectives up to the endpoint of all major foreseeable outcomes. In other words, we imagine the moral reaction to our decision in every eventuality. By hypothetically retrospecting on potential branches of future development, we can build ethical arguments about what to do in the present.

\citeauthor{kolker2023uncertain} adapted this procedure into \textit{Machine Ethics Hypothetical Retrospection} (MEHR)~\cite{kolker2023uncertain}, described in Figure~\ref{fig:MEHR example}.
In summary, an \textit{Ethical Decision Problem} contains a set of states, actions and potential branches of future development.
Based on a Value-Based Argumentation Framework~\cite{atkinson2016value,dung1995acceptability}, arguments in favour of each action are generated from the perspective of their branches of development using an \emph{argument scheme}:

\begin{quote}
    \emph{``From the initial state $I$, it was acceptable to perform action $\mathbf{a}$, resulting in consequences $\mathbf{b}$ with probability $\mathbb{P}$."}    
\end{quote}

Attacks are generated between arguments in support of different actions by two \emph{critical questions}. 
For every moral theory under consideration, the argument from branch $\mathbf{b_1}$ in support of action $\mathbf{a_1}$ attacks the argument from branch $\mathbf{b_2}$ in support of action $\mathbf{a_2}$ when the following are affirmative:

\begin{quote}
    \noindent \textbf{CQ1:}~~Did $\mathbf{b_2}$ violate a moral theory that $\mathbf{b_1}$ did not?
    
    \noindent \textbf{CQ2:}~Did $\mathbf{a_2}$ hold a greater probability of violating the moral theory than $\mathbf{a_1}$?    
\end{quote}

Here, the moral theory is a free variable, intentionally non-specific to generalise across theories. MEHR has been demonstrated with Utilitarianism, assigning numeric values to outcomes and Deontology, assigning binary duties to actions. Other theories should be compatible~\cite{kolker2023uncertain}.
Argument attacks can be generated from multiple, potentially conflicting moral theories, meaning MEHR extends Hansson's procedure for problems with moral uncertainty.
When an argument is attacked, it represents negative retrospection: another action could have avoided the violation and this was foreseeable. This indicates the decision-making was flawed and so there is negative retrospection.
Ideally, an action is selected with no negative retrospection on any of its potential outcomes. If no such action exists, the action with minimal probability of resulting in negative retrospection is selected. This probability is referred to as \emph{non-acceptability}.
We note that MEHR simplifies hypothetical retrospection by only considering the perspective of outcome endpoints without intermediate perspectives. 
\section{Related Work}
There are different approaches to Machine Ethics planning.

Often, a quantitative approach is used with Act-Utilitarianism~\cite{sep-utilitarianism-history} and decision theory, i.e., to maximise expected utility. 
The Utilibot Project is an architecture for autonomous robots in the home~\cite{cloosUtilibot}.
It integrates Bayesian Networks representing uncertain knowledge with MDP actions and utilities, using Policy Iteration to maximise expected utility.
Utilitarianism as a sole decision mechanism is problematic. It assumes all outcomes can be appraised in terms of a single number\footnote{An outcome where a person is injured may have \texttt{-10} utility and an outcome where they lose their life may have \texttt{-100} utility. Thus, \texttt{10} injuries is equal to a loss of life, which does not match intuition.} without a clear or agreed methodology. \citeauthor{brundage2014limitations} discusses other critiques of Consequentialism in~\cite{brundage2014limitations}.

Alternatively, there are qualitative approaches. \citeauthor{jedwabny2021generating}~\cite{jedwabny2021generating} create a linearization from a ranking over ethical rules.
For cognitive agents, \citeauthor{dennis2014ethical} derive a substantive order over plans from a ranking of formal ethical principles~\cite{dennis2014ethical}.
\citeauthor{lindner2017hera} develop techniques to evaluate plans using ethical theories~\cite{lindner2017hera,lindner2020evaluation}; \citeauthor{Dennis_Bentzen_Lindner_Fisher_2021} apply this to cognitive agents~\cite{Dennis_Bentzen_Lindner_Fisher_2021}, 
However these approaches assume determinism.
\citeauthor{grandi2023moral} evaluate plans in a Multi Agent System with ranked logical formulae. Uncertainty over other agents' actions is handled optimistically, pessimistically, or to avoid blame~\cite{grandi2023moral}, all without probability as in our approach.

\citeauthor{rodriguez2020structural} use a Multi-Agent MDP with norms and an action praiseworthiness function integrated with existing, non-moral rewards. They adjust penalties and praiseworthiness to ensure agents learn ethically aligned policies in Reinforcement Learning~\cite{rodriguez2020structural}. Outcome uncertainty is handled with utility expectation.

Similarly, the Ethically Compliant Autonomous System (ECAS) framework represents MDPs with a non-moral reward function and moral principles~\cite{svegliato2021ethically,nashed2021ethically}. Agents have conflicting value functions which determine the truth of a moral principle function. The MDP is solved as a linear program, maximising expected reward subject to a single moral principle constraint.
ECAS evaluates policies individually and there may be no moral solutions, unlike MEHR which finds the best available policy with minimal non-acceptability.

For problems with moral uncertainty between moral theories, \citeauthor{ecoffet2021reinforcement} apply a voting system in Reinforcement Learning~\cite{ecoffet2021reinforcement}. Moral theories are trained to effectively express their views at each time step using a limited voting budget.
Preference between theories is represented by some credence  $c\in (0,1]$ instead of the weak lexicographic order in our paper.
Theory preferences are expressed by utilities only and outcome uncertainty by expectation.

\citeauthor{stojanovski2024ethical} define an MDP with positive/negative values that state-action transitions can promote/demote and repair (treating previous actions). They build an ethical reward function that lexicographically avoids harm then, then maximises good~\cite{stojanovski2024ethical}.
They suggest a hierarchy over MDP values and relative strength in promotion/demotion for future work.
\section{Multi-Moral Markov Decision Process}
\label{sec:MMMDP}
We model our planning problem as a \textit{Multi-Moral Markov Decision Process} (MMMDP) based on a MOMDP.
\begin{definition}[Multi-Moral Markov Decision Process]
    \leavevmode\\
    A tuple $\langle S, A, P, s_0, H, M, C, L\rangle$ with a finite set of states $S$, a finite set of actions $A$ and a probabilistic transition function $\probf: S \times A \times S \rightarrow [0,1]$.
    There is an initial state $s_0 \in S$ and $H\in \mathbb{N}$ is the finite horizon.
    
    $M$ is a finite set of moral Theories. $C$ is a finite set of \emph{moral considerations} and $L : M \rightarrow \mathbb{Q}$ is a weak lexicographic ranking.
\end{definition}

States and dynamics are identical to ordinary finite-horizon MDPs (see Section~\ref{sec:background}).
Briefly, moral theories in $M$ are internally consistent moral perspectives, such as Act-Utilitarianism or Deontology~\cite{bench2020ethical}. They define an attack relation for MEHR argumentation, detailed in Section~\ref{sec:MEHR}.
Judgements are based on some moral considerations in $C$, such as utility functions or duties.
Preferences between theories are defined with the weak lexicographic order $L$.
In the following two sections we will explain each component in detail with a running example of The Lost Insulin.


\subsection*{States, Actions and Dynamics}
We use the ethical dilemma discussed by \citeauthor{coleman1992risks}~\cite{coleman1992risks}, adapted to Machine Ethics by \citeauthor{atkinson2008addressing}~\cite{atkinson2008addressing}.
Hal is a diabetic who, through no fault of his own, has lost his insulin supply and needs some urgently to stay alive.
He knows his neighbour Carla has some, but does not have permission to enter her house and take it.
Is Hal justified in stealing the insulin to save his life?

\begin{figure}
    \centering
    \usetikzlibrary{positioning, arrows.meta}

\newcommand{\widthFactor}{1.5}
\newcommand{\heightFactor}{1.2}

\tikzset{scr/.style={-{Triangle[width=1mm]}}}
\begin{tikzpicture}[
    sumnode/.style={rectangle, draw, align=center, minimum width=0.5cm, minimum height=0.5cm, font=\huge},
    actionnode/.style={trapezium, draw, trapezium angle=70, trapezium stretches=true, trapezium stretches body=true, align=center, font=\huge},
    level 1/.style={level distance=\heightFactor * 1cm, sibling distance=\widthFactor * 1.8cm},
    level 2/.style={level distance=\heightFactor * 0.9cm, sibling distance=\widthFactor * 0.7cm},
    probnode/.style={draw, circle, align=center, inner sep=0.01cm, font=\large},
    attack/.style={line width = 1pt, -{Latex[width=3mm]}},
    argument/.style={draw, rectangle, align=center},
  utilNonAcc/.style={color=red, align=right},
  deonNonAcc/.style={color=blue, align=right, text width=1cm},
  attack/.style={->, -{Stealth[width=2mm]}, line width=0.2mm}
]

\node[sumnode] at (0,0) (root) {$s_0$}
child[scr] { 
    node[actionnode] (a1) {$a_1$}
    child[-] {
        node[probnode] {$\ProbSpAqSq$}
        child[scr] {
            node[sumnode] (s5) {$s_1$}
        }
    }
}
child[scr] {
    node[actionnode] (a2) {$a_2$}
    child[-] { 
        node[probnode] {$\ProbSpAwSw$}
        child[scr] {
            node[sumnode] (s5) {$s_2$}
        }
    }
    child[-] { 
        node[probnode] {$\ProbSpAwSe$}
        child[scr] {
            node[sumnode] (s5) {$s_3$}
        }
    }
    child[-] { 
        node[probnode] {$\ProbSpAwSr$}
        child[scr] {
            node[sumnode] (s5) {$s_4$}
        }
    }
    child[-] { 
        node[probnode] {$\ProbSpAwSt$}
        child[scr] {
            node[sumnode] (s5) {$s_5$}
        }
    }
};

\node[probnode] (s0a1s0Prob) at (-1.7 * \widthFactor, -0.3 * \heightFactor) {$\ProbSpAqSp$};

\draw[-] (a1.south) to[out=250, in=230, looseness=1.5] (s0a1s0Prob);
\draw[scr] (s0a1s0Prob) to[out=70, in=180, looseness=0.5] (root);

\end{tikzpicture}
    \caption{In $s_0$ Hal has no insulin; $s_1$ Hal dies waiting; $s_2$ both live; $s_3$ Hal dies; $s_4$ Carla dies; $s_5$ both die.}
    \label{fig:insulin_mdp}
\end{figure}
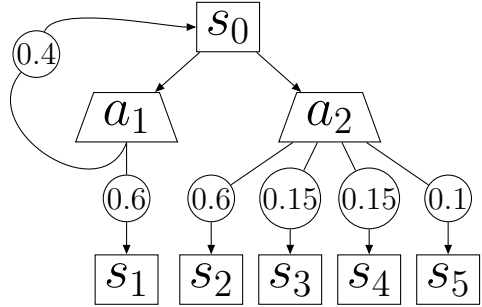

We model the problem as a MMMDP visualised in Figure~\ref{fig:insulin_mdp}.
In the initial state $s_0$ Hal has no insulin. With action $a_1 \in A$ Hal waits and there is a $P(s_0,a_1,s_1)=\ProbSpAqSq$ probability of transitioning to $s_1$ where Hal dies; there is $P(s_0,a_1,s_0)=\ProbSpAqSp$ probability that Hal lives and faces the choice again.
In action $a_2 \in A$, Hal goes to Carla's house and takes the insulin. In state $s_2$, Carla does not need the insulin and they both live at probability \ProbSpAwSw; in state $s_3$, the stolen insulin is insufficient and Hal dies with probability \ProbSpAwSe; in state $s_4$, Carla dies without her insulin at probability \ProbSpAwSr; finally, in state $s_5$ they both die with probability $\ProbSpAwSt$.
We let the horizon $H=2$ represent the hours until Hal receives his regular insulin delivery and the dilemma no longer applies.
If Hal dies before the horizon, we model it with repeated wait actions $a_0$.
Thus, while there is a $P(s_0,a_0,s_0)=\ProbSpAqSp$ chance Hal survives by waiting through the first hour, the chance Hal lives until his insulin resupply is $\ProbSpAqSp^2=0.16$. 

\subsection*{A Moral Consideration}
Meaning is extracted from the problem using moral considerations $C$ which track sources of morally relevant information or moral worth.
We associate states in the MMMDP with worth using a \emph{worth function} $W_c : S \rightarrow \estimateType_c$.
For notation, we may interpret the worth function as a sequence $W(s_i)=w_i$ for $s_i \in S$.
We define a moral consideration formally below.

\begin{definition}[Moral Consideration]\label{def:moral_theory}
    \leavevmode\\
    A Moral Consideration is a tuple $c= \moralTheory$.
    \begin{itemize}
        \item $\estimateType$ is a type representing the space of information required for decisions. They are referred to as moral worth.
        \item $\judgef : S \times A \times S \rightarrow \estimateType$ is a judgement function that extracts the morally relevant information from a state-action transition.
        \item $\gatherBasef^{W} : (\estimateType \times [0,1])^n \rightarrow \estimateType$ 
        aggregates morally relevant information with probabilities into a single estimate of future moral worth. It uses a baseline of  existing morally relevant information from a \emph{worth function} $W: S \rightarrow \estimateType$, subject to \emph{Assumption \ref{asm:additive}} (additivity over moral worth).
        \item \estGEQ is a transitive preference relation over $\estimateType$.
        \item $\approx : \estimateType \times \estimateType \rightarrow \{\top, \bot\}$ maps to $\top$ when moral worth is \textit{consistent}, $\bot$ otherwise.
    \end{itemize}   
    \label{def:moralTheory}
\end{definition}

A moral consideration represents a source of morally relevant information in the planning domain.
It may be considered a generalisation of an MDP's reward function.
This is clearest when formalising the moral consideration of utility associated with Act-Utilitarianism:
$\estimateType$ becomes the space of real numbers $\mathbb{R}$ representing utility gained/lost through state transitions $\judgef(s,a,s')$.
The aggregation function $\gatherBasef^W(W', P) = \sum_{i \in 1...|W'|} P_i \cdot (w'_i + w_i)$ calculates expected utility where $W'$ is a set of state utilities, $P$ are associated probabilities, and $W$ are baseline utilities\footnote{We assume for simplicity that $W$, $P$ and $W'$ have equal magnitude.}.
Relation $\succeq$ is the standard for $\mathbb{R}$. Consistency relation $w \approx w'$ is $\top$ when $|w' - w| < \epsilon$ for some small constant $\epsilon>0$, $\bot$ otherwise.

For the Lost Insulin, suppose Act-Utilitarianism appraises death with $-10$ utility, i.e., $\judgef(s_0,a_1,s_0)=0$, $\judgef(s_0,a_1,s_1)=-10$, $\judgef(s_0,a_2,s_5)=-20$.
The expected utility of stealing ($a_2$) at $s_0$ is expressed by aggregation function $\gatherBasef^W(\{ \judgef(s_0,a_2,s') : \forall_{s'\in S}\}, \{P(s_0,a_2,s') : \forall_{s'\in S}\})$.
This is
$\ProbSpAwSw(0+W(s_2))+\ProbSpAwSe(-10+W(s_3))+\ProbSpAwSr(-10+W(s_4))+\ProbSpAwSt(-20+W(s_5))$
and with baseline $W$ initialised to 0, this is $\ProbSpAwSw(0+0)+\ProbSpAwSe(-10+0)+\ProbSpAwSr(-10+0)+\ProbSpAwSt(-20+0)=-5$ utility\footnote{Successor states from $a_2$ with $0$ probability are omitted from the equations for space.}.

For syntactic sugar, we abstract state-action aggregation as follows:
$\gatherf^W(s,a)= \gatherBasef^W(\{ \judgef(s,a,s') : \forall_{s'\in S}\}, \{P(s,a,s') : \forall_{s'\in S}\})$.
And so, $\gatherf^W(s_0,a_2)=-5$.
For reasons in Section \ref{sec:MEHR}, alternate states and probabilities may be passed as follows
$$\gatherf^W(s,a,S',P')= \gatherBasef^{W'}(\{\judgef(s,a,s') : \forall_{s' \in S'} \}, \{P'(s,a,s') : \forall_{s'\in S'}\})$$
where $W'$ contains worth for states in $S'$: $W' = \{W(s) : s \in S' \}$.

We are interested in non-stationary, deterministic policies $\policy : S \times \{0,\ldots,H-1\} \rightarrow A$. We exclude stochastic policies for simplicity and because, arguably, randomness is not acceptable in ethical decision making.
A moral consideration evaluates a policy with respect to every state iteratively using a \textit{non-stationary worth function}.
We interpret this as an $H+1 \times |S|$ matrix where $W[t]$ is a worth function for all states at time step $t\in \{0,...,H\}$; $W[t](s)$ is the expected worth at time $t$, state $s$. The worth function is defined as
$$W^\policy[t](s)=\gatherf^{W^\policy[t+1]}(s,\policy(s,t)), \forall{s \in S, t \in \{0,\ldots,H-1\}}$$
The vector $W[H]$ contains null values in $\estimateType$ so as to not affect aggregation in $\gatherBasef$.
One worth function is preferred to another if it is preferable at every state and time.
A \textit{consideration-optimal worth function} is the most preferable worth achievable at each state and time. This is calculated like the \citeauthor{Bellman_DynamicProgramming_57} Optimality Equation~\cite{Bellman_DynamicProgramming_57}: $$W^*[t](s)=\max^\succeq_{a\in A} \gatherf^{W^*[t+1]}(s,a), \forall{s \in S, t \in \{0,\ldots,H-1\}}$$
This assumes moral worth is additive through $\gatherBasef$.
\begin{assumption}[Additive Moral Worth]
    \label{asm:additive}
    Let $W'^+, W'^- \in \estimateType^n$ be worth functions, for any $n$, and let $P'^+, P'^-$ be sets of probabilities such that $(W'^+, P'^+)$ \textbf{is preferred to} $(W'^-, P'^-)$, i.e., $\gatherBasef^{\estf}(\estf'^+, P'^+) \prec \gatherBasef^{\estf}(\estf'^-, P'^-)$ for any baseline $\estf \in \estimateType$.

    Furthermore, let $\estf^{+}, \estf^{-} \in \estimateType^n$ be worth functions such that $\estf^{+}$ \textbf{is preferred to} $\estf^{-}$, i.e., $\gatherBasef^{\estf^{+}}(W',P') \prec \gatherBasef^{\estf^{-}}(W',P')$ for any worth function $W' \in \estimateType^n$ and any probabilities $P' \in [0,1]^n$.

    It must be the case that $\gatherBasef^{\estf^{-}}(\estf'^-, P'^-) \prec \gatherBasef^{\estf^{+}}(\estf'^+, P'^+)$.
\end{assumption}


\subsection*{Many Moral Considerations}
MMMDPs may have plenty of moral considerations.
Suppose an absolute moral theory where any probability of breaking the law is wrong.
A moral consideration's worth could be represented by $\estimateType=\{\top, \bot\}$ where $\top$ violates the law and $\top \prec \bot$.
In the example, $\judgef(s_0,a_2,s_2)=\top$ and $\judgef(s_0,a_1,s_1)=\bot$.
Worth $\estimateType$ aggregates with $\gatherBasef^W(W', P) = \bigvee_{i \in |W'|} (P_i>0 \land w'_i) \lor w_i$. Worth is consistent only if it is equal, $w \approx w' = w == w'$.
Thus, if $W[t](s)=\bot$ for all $t \in \{0,\ldots,H\}$ and $s\in S$, then $\gatherf^{W}(s_0,a_1)=\gatherBasef^W((\bot \lor \bot) \lor \ldots \lor (\bot \lor \bot))=\bot$ (no law violated) and $\gatherf^{W}(s_0,a_2)=\top$ (law violated).

For notation, we distinguish moral considerations with subscripts: consideration $c_i \in C$ is the tuple $\langle \estimateType_j, \judgef_i, \gatherBasef^W_i, \preceq_i, \approx_i \rangle$.
To hold the worth for all considerations in $C$, we use a \emph{multi-worth function} $\boldsymbol{W}=\{W_1, \ldots, W_{|C|}\}$ where $W_i\in \boldsymbol{W}$ is a worth function for consideration $c_i\in C$.
We hold a single element of worth from all moral considerations using a \emph{worth-vector} $\vec{w} \in 2^{\{\estimateType_1,\ldots,\estimateType_{|C|}\}}$.
We generalise $\gatherf$ for multiple considerations as $\vgatherf$, which returns expectations as a worth-vector.
We bend notation to apply the worth-vector across a multi-worth function at state $s$ and time $t$: $\boldsymbol{W}^\policy[t](s) = \vgatherf^{\policy} (s,\policy(s,t))= \{\gatherf_i^\policy(s,\policy(s,t)) : \forall{i \in 1,...,|C|}\}$. We demonstrate in Figure~\ref{fig:notation}.

\begin{figure}
    \centering
    \usetikzlibrary{positioning, arrows.meta}

\begin{tikzpicture}[]

\newcommand{\widthFactor}{0.4}
\newcommand{\heightFactor}{0.2}

\setlength{\arraycolsep}{2pt} 

\node[] (deon-worthFunction) {
$W_{abs} = \begin{bmatrix}
\top & \bot & \bot\\
\top & \top & \bot\\
\top & \bot & \bot
\end{bmatrix}$};
\node[below=\heightFactor * 0.6cm of deon-worthFunction] (deon-worthfTime) {$W_{abs}[1]=(\bot,\top,\bot)$};
\node[below=\heightFactor * 0.6cm of deon-worthfTime] (deon-worthfTimeState) {$W_{abs}[1](s_1)=\top$};

\node[right=\widthFactor * 0.3cm of deon-worthFunction] (util-worthFunction) {
$W_{util} = \begin{bmatrix}
1 & 2 & 0\\
3 & 4 & 0\\
5 & 6 & 0
\end{bmatrix}$};
\node[above left=\heightFactor * -0.9cm and \widthFactor * -6.6cm of util-worthFunction] (util-timeLabel) {$\mathit{t=}\begin{matrix}
\mathit{0} & \mathit{1} & \mathit{2}
\end{matrix}$};

\node[draw=none, right=\heightFactor * -0.7cm of util-worthFunction, align=left] (util-stateLabel) {$\mathit{s_0}$\\$\mathit{s_1}$\\$\mathit{s_2}$};

\node[below=\heightFactor * 0.6cm of util-worthFunction] (util-worthfTime) {$W_{util}[1]=(2,4,6)$};
\node[below=\heightFactor * 0.6cm of util-worthfTime] (util-worthfTimeState) {$W_{util}[1](s_1)=4$};

\node[above right=\heightFactor * -1cm and \widthFactor * 1.5cm of util-worthFunction] (multiworthf) {$\boldsymbol{W}=\{W_{abs}, W_{util}\}$};

\node[below=\heightFactor * 0.4cm of multiworthf] (multiworthfTimeState) {$\boldsymbol{W}[1](s_0)=\vec{w}=\{\top, 4\}$};

\node[align=left, below=\heightFactor * 1cm of multiworthfTimeState] (m) {$\begin{bmatrix}
\{\{\top,1\}\} & \ldots & \{\{\bot,0\}\}\\
\{\{\top,3\}\} & \ldots & \{\{\bot,0\}\}\\
\{\{\bot,5\}\} & \ldots & \{\{\bot,0\}\}
\end{bmatrix}$};
\node[above=\heightFactor * -0.9cm of m] {$\vec{W} =$};

\node[align=left, below=\heightFactor * 0.2cm of m] (mn) {$\vec{W}=\{\vec{w} \ldots \}$};

\end{tikzpicture}
    \caption{Shows notation. Worth function $W$ maps state-time to worth. Multi-worth function $\boldsymbol{W}$ holds many worth functions. Worth-vectors $\vec{w}$ store worth across considerations. From Section~\ref{sec:algorithm}, $\vec{W}$ maps state-time to a set of worth-vectors.}
    \label{fig:notation}
\end{figure}

\subsection*{Moral Theories}
A moral theory is a domain-dependent representation of an ethical perspective.
It defines the nature of hypothetical retrospection based on worth from moral considerations; a moral theory decides how policies and branches of development are compared.
\begin{definition}[Moral Theory]
    \leavevmode\\
    A moral theory $m\in M$ is a tuple $m=\langle C^m, \CQ \rangle$ where $C^m \subseteq C$ is a subset of moral considerations from a MMMDP.
    Function $\CQ : 2^{Arg} \rightarrow \{\attack, \circ\}$ determines if \textit{arguments} should attack each other according to this theory (see Section~\ref{sec:MEHR}).
\end{definition}
In this paper, we only consider moral theories with a single moral consideration, though the formalism supports many. In that case, the attack relation would define how considerations are balanced.

Moral theories may conflict; stakeholders may believe some theories more than others. This is expressed with a weak lexicographic order.
For theories $m, m' \in M$, if $L(m)<L(m')$, any worth from consideration $C^m$ is prioritised over worth from $C^{m'}$.
In the running example, a Utilitiarian-optimal worth function has $W^*_{util}[0](s_0)=-5$ by stealing insulin. However, the absolute optimal worth is $W^*_{abs}[0](s_0)=\bot$ by waiting.
The same policy cannot optimise both.
If $L(m_{util}) < L(m_{abs})$, the Utilitarian's optimal policy would be preferable; on the other hand if $L(m_{util}) > L(m_{abs})$, then the absolute no-stealing policy is preferred. If $L(m_{util}) = L(m_{abs})$, the stakeholder has no preference or moral uncertainty. Another mechanism is required to differentiate the policies such as MEHR.
\section{Machine Ethics Hypothetical Retrospection over Policies}
\label{sec:MEHR}
Machine Ethics Hypothetical Retrospection (MEHR)~\cite{kolker2023uncertain} is an argumentation procedure~\cite{dung1995acceptability} for Machine Ethics decision making with outcome and moral uncertainty.
We contribute a novel formalism for probabilistic planning.
It creates an argument in favour of each policy from the perspective their potential branches for future development and models negative retrospection as argument attacks.
We model a branch of development as a history/trajectory in the MMMDP $h=\langle S^h,A^h \rangle$ where $S^h$ is a sequence of states and $A^h$ is a sequence of actions.
$S^h_1$ is the start state, action $A^h_1$ transitions to state $S^h_2$ etc., i.e.,
$h \equiv \{S^h_1, A^h_1, S^h_2, A^h_1, \ldots, S^h_{H+1}\}.$
The probability of a history is $P(h) = \prod^{|A^h|}_{i=1} P(S^h_i, A^h_i, S^h_{i+1})$.
A history's worth is found using the same aggregation function as with policies:
$$W^h[t](s_i) = 
\gatherf^{W^h[t+1]}(s_i, a_i, \{s_{i+1}\}, \{1.0\})
$$
A single successor state is specified and 1.0 probability is passed to find a history's worth.
For a policy $\policy$, the worth and probability of all its histories can be found using Depth-First-Search.
The algorithm exploits Assumption \ref{asm:additive} by aggregating worth while traversing from $s_0$ to the horizon. For transparency, there is pseudocode in Algorithm~\ref{alg:getOutcomeWorths}; we do not examine it since we do not consider it a notable contribution.
\begin{algorithm}[t]
\begin{minipage}[t]{\textwidth}
  \raggedright
  \textbf{Input:} Policy $\pi$, multi-worth function $\boldsymbol{W}$\\
  \textbf{Output:} Mapping $\phi : \{\mathcal{W}_c : \forall{c\in C}\} \rightarrow [0,1]$
\end{minipage}
\begin{algorithmic}[1]
\Procedure{ExtractHistories}{$s$, $t$, $\vec{w}$, $pr$, $\phi$}
    \If{$t \geq H \text{ \textbf{or} } \{s'\in S : P(s,a,s')>0\} \textbf{ is } \emptyset$}
        \State $\phi(\vec{w}) \gets \phi(\vec{w}) + pr$
        \State \Return
    \EndIf
    \For{$\{s' \in S, a \in A : P(s,a,s')>0\}$}
        \State $\boldsymbol{W}' \gets \{\vec{w}_c : c\in C\}$ \text{// Set baseline to prev. worth}
        \State $\vec{w}' \gets 
\gatherf^{\boldsymbol{W}'}(s, a, \{s'\}, \{1.0\})$
        \State $ExtractHistories(s', t+1, \vec{w}', pr \cdot P(s,a,s'), \phi)$
    \EndFor
\EndProcedure
\State $\phi(\vec{w}) \gets 0 : \forall{\vec{w} \in 2^{\{\mathcal{W}_c : \forall{c\in C}\}}} \text{// Set default to map 0. }$
\State $ExtractHistories(s_0, 0, \{\text{null } \mathcal{W}_{c} : \forall_{c\in C}\}, 1.0, \phi)$
\State \Return $\phi$
\end{algorithmic}
\caption{Extract history worth and probabilities from policy}
\label{alg:getOutcomeWorths}
\end{algorithm}
For the Lost Insulin running example, the `waiting policy' always selects $a_1$ and has three potential histories: $h_1\equiv \{s_0, a_0, s_0, a_0, s_0\}$ (waiting two hours), $h_2\equiv \{s_0, a_0, s_0, a_0, s_1\}$ (waiting two hours, then collapse), $h_3\equiv \{s_0, a_0, s_1, a_0, s_1\}$ (waiting one hour, then collapse).

We take the expected worth of a policy as its worth at the initial state $s_0$ and time $t=0$, $\vgatherf^\policy(s_0,\policy(s_0,0))$.
Thus, from a selection of alternate policies, we use MEHR argumentation to select an ethical policy~\cite{kolker2023uncertain}.
The procedure translates naturally. An argument $Arg$ is generated from the perspective of each history endpoint in support of the preceding policy.
Arguments are generated by an \textit{Argument Scheme}.
We define $Arg(\policy, h)$ as
\begin{quote}
    ``From the initial state $s_0$, it was right to use policy $\policy$, resulting in the history $h=\{S^h_1,A^h_1 ... A^h_{H-1}S^h_H\}$ with probability $P(h)$.''
\end{quote}
This may be interpreted as the default hypothetical retrospective argument, though it may be invalidated by other retrospections.
There may be negative retrospection for a policy after an outcome if an alternate policy is preferred.
We define this as an attack relation $\CQ$ for each moral theory $m \in M$.
An argument $Arg(\policy, h)$, supporting policy $\policy$ from history $h$, attacks an opposing argument $Arg(\policy', h')$ if it causes negative retrospection: reveals a decision making flaw in choosing $\policy'$.
For each moral theory, we define the attack relation in terms of Critical Questions~\cite{atkinson2008addressing, kolker2023uncertain}.
There is an attack when $\CQ(Arg(\policy, h), Arg(\policy', h'))=\attack$ and the following are affirmative:
\begin{quote}
    \textbf{CQ1:}~~Did $h'$ violate the moral theory and $h$ did not?\\
    \textbf{CQ2:}~~Was there greater expectation that $\policy'$ would violate the moral theory than $\policy$?
\end{quote}
This represents that an alternative history was preferable and expected to be preferable, resulting in negative retrospection and undermining the target argument.
Each moral theory determines the meaning of a violation and violation expectation.
For Act-Utilitarianism, the questions are
\textbf{CQ1:}~~Did $h$ have greater utility than $h'$?
\textbf{CQ2:}~~Did $\policy$ have greater expected utility than $\policy'$?
For each theory the attack is defined as
    $\CQ(Arg(\policy, h), Arg(\policy', h')) = \attack \textbf{ if }CQ1 \land CQ2 \textbf{ otherwise } \circ$
when
\begin{align*}
CQ1 = W^h[0](s_0) \succ W^{h'}[0](s_0)\\
CQ2 = \gatherf^\policy(s_0, \policy(s_0, 0)) \succ \gatherf^{\policy'}(s_0, \policy'(s_0, 0))
\end{align*}

Thus, there is a supportive argument generated from the outcome of every policy; moral theories suggest attacks between arguments that support different policies.
For each argument, we define its attackers as a set of argument and moral theory pairs:
\begin{align*}
  \mathcal{X}(Arg(\policy, h)) =\\ \{(m, Arg(\policy', h')) : \forall{m\in M},\\
  \CQ_m(Arg(\policy', h'), Arg(\policy, h)) = \attack \land \not\exists m' \in M, L(m') < L(m) \land \\\gatherf^{W^{\policy'}[1]}_{m'}(s_0 ,\policy'(s_0 ,0)) \prec \gatherf^{W^{\policy}[1]}_{m'}(s_0 ,\policy(s_0 ,0))\}
\end{align*}
In words, there is a potential attack on an argument for every other policy $\policy'$, each of their histories $h'$, and every moral theory $m$.
MEHR attacks are based the moral theories' attack relation $\CQ$, but an attack is blocked if there is a lexicographic preferred theory such that $\policy$ is preferred.
Note, there can be multiple attacks from one argument to another, unlike traditional argumentation~\cite{dung1995acceptability}.

The non-acceptability of a policy is its probability of reaching an outcome with negative retrospection for each moral theory. Equivalently, it is the number of attacks on each supporting argument multiplied by the argument history's probability:
$$\mathcal{N}(\policy)= \sum_{h\text{ from } \policy}{P(h) \cdot |\mathcal{X}(Arg(\policy, h))|}$$
A policy with 0 non-acceptability should be selected if one exists. Otherwise, there is a `generalised case of a moral dilemma'~\cite{hansson2013ethics} and we choose the policy with minimal non-acceptability instead.
We define the \emph{morally acceptable policies} as those minimising non-acceptability $$\Pi^* = \{\policy : \mathcal{N}(\policy) = \min_{\policy}\mathcal{N}(\policy)\}.$$ We visualise MEHR with The Lost Insulin example in Figure 4.

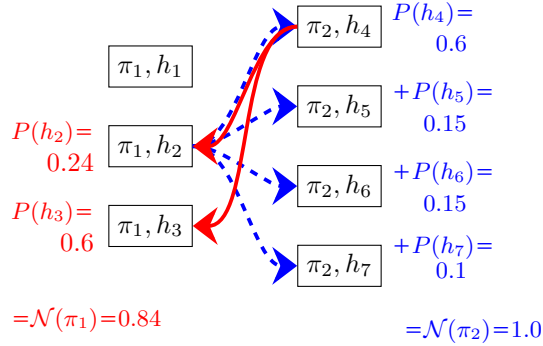
\begin{figure}
    \centering
    \usetikzlibrary{positioning, arrows.meta, calc}
\begin{tikzpicture}[
  grow=down,
  outcome/.style={circle,draw, minimum size=1cm, align=center, text width=1cm, font=\footnotesize}, 
  parallelogram/.style={draw, trapezium, trapezium angle=70, minimum height=1cm,  trapezium stretches=true, trapezium stretches body=true, align=center, text width=1cm, font=\footnotesize}, 
  attack/.style={line width = 1pt, -{Latex[width=3mm]}},
  argument/.style={draw, rectangle, align=center},
  utilNonAcc/.style={color=red, align=right},
  deonNonAcc/.style={color=blue, align=right, text width=1cm},
  attack/.style={->, -{Stealth[width=5mm]}, line width=0.5mm}
]


\newcommand{\widthFactor}{1.3}
\newcommand{\heightFactor}{0.5}

\node[argument] (pi2_h4) {$\policy_2, h_4$} ;
\node[argument, below=\heightFactor * 1cm of pi2_h4] (pi2_h5) {$\policy_2, h_5$};
\node[argument, below=\heightFactor * 1cm of pi2_h5] (pi2_h6) {$\policy_2, h_6$};
\node[argument, below=\heightFactor * 1cm of pi2_h6] (pi2_h7) {$\policy_2, h_7$};

\node[argument, left=\widthFactor * 1.5cm of $(pi2_h4)!0.5!(pi2_h5)$] (pi1_h1) {$\policy_1, h_1$};
\node[argument, below=\heightFactor * 1cm of pi1_h1] (pi1_h2) {$\policy_1, h_2$};
\node[argument, below=\heightFactor * 1cm of pi1_h2] (pi1_h3) {$\policy_1, h_3$};

%
%
\draw[attack, color=blue, dashed] (pi1_h2.east) to[out=0, in=180, looseness=0.7] (pi2_h4.west);
\draw[attack, color=blue, dashed] (pi1_h2.east) to[out=0, in=180, looseness=0.7] (pi2_h5.west);
\draw[attack, color=blue, dashed] (pi1_h2.east) to[out=0, in=180, looseness=0.7] (pi2_h6.west);
\draw[attack, color=blue, dashed] (pi1_h2.east) to[out=0, in=180, looseness=0.7] (pi2_h7.west);

\draw[attack, color=red] (pi2_h4.west) to[in=0, out=180, looseness=0.7] (pi1_h2.east);
\draw[attack, color=red] (pi2_h4.west) to[in=0, out=180, looseness=0.7] (pi1_h3.east);

\node[utilNonAcc, align=right, left=\widthFactor * 0.05cm of pi1_h2] (n) {\small$P(h_2)$=\\$0.24$};
\node[utilNonAcc, align=right, left=\widthFactor * 0.05cm of pi1_h3] {\small$P(h_3)$=\\$0.6$};

\node[deonNonAcc, right=\widthFactor * 0.01cm of pi2_h4] (m) {\small$P(h_4)$=\\$0.6$};
\node[deonNonAcc, right=\widthFactor * 0.01cm of pi2_h5] {\small +$P(h_5)$=\\$0.15$};
\node[deonNonAcc, right=\widthFactor * 0.01cm of pi2_h6] {\small+$P(h_6)$=\\$0.15$};
\node[deonNonAcc, right=\widthFactor * 0.01cm of pi2_h7] (pi2_h6_prob){\begin{spacing}{0.85}\small+$P(h_7)$=\\$0.1$\end{spacing}};

\node[color=blue, anchor=east] (non_accept_pi2) at ([xshift=1cm, yshift=-\heightFactor * 0.5cm] pi2_h6_prob.south -| m.east) {\small=$\mathcal{N}(\pi_2)$=$1.0$};

\node[color=red, anchor=west] (non_accept_pi1) at ([yshift=-\heightFactor * 0.2cm] pi2_h6_prob.south -| n.west) {\small=$\mathcal{N}(\pi_1)$=$0.84$};


\end{tikzpicture}
    \label{fig:worked}
    \caption{MEHR argumentation graph. Boxes are arguments from policy-history pairs, $Arg(\pi,h)$. Directed edges are attacks, meaning $\Psi_m(Arg(\pi', h'), Arg(\pi', h'))$. Red/solid edge for utility attack; blue/dashed line for no-stealing attack. Probabilities on attacked arguments sum to policy's non-acceptability $\mathcal{N}(\policy)$.}
\end{figure}

We note this formalism's philosophical assumptions. First, that an action's moral worth can be aggregated and combined with probabilities--that an aggregation function $\gatherf$ can be expressed for moral theories. As in MEHR, we assume the perspective of the endpoint of action outcomes is sufficient for moral decision-making. \citeauthor{hansson2013ethics}'s original Hypothetical Retrospection argues relevant information can be described in terms of consequences, though \citeauthor{hansson2013ethics} considers all perspectives up to an action's endpoint, rather than the endpoint perspective alone. MEHR also differs by allowing conflicting moral theories under moral uncertainty.

\section{A Non-moral Extension}
We make our formalism compatible with non-moral costs and goals using ideas from the Stochastic Shortest Path (SSP) Problem~\cite{Bellman_DynamicProgramming_57}.
\begin{definition}[Multi-Moral Stochastic Shortest Path Problem]
  \leavevmode\\
    A Multi-Moral Stochastic Shortest Path Problem (MMSSP) is a tuple \\ $\langle S, A, P, s_0, H, M, C, L, R, \budget, \goals \rangle$ with components identical to a MMMDP.
    Additionally: a non-moral cost consideration $R \in C$, defined as a special type of moral consideration; a positive \emph{budget} $\budget \in \mathbb{R}^+$; and a set of goal states $\goals \subseteq S$ that are absorbing: $P(s,a,s')=0, \forall{s \in G, a\in A, s' \in S \backslash G}$ and $P(s,a,s)=1, \forall{s \in G, a\in A}$.
\end{definition}

A MMSSP has competing moral theories and quantified costs over state transitions.
For simplicity, costs are a moral consideration, almost identical to the utility consideration discussed in Section~\ref{sec:MMMDP}.
This consideration is not associated with any moral theory.
Also, state transitions must cost more than 0, i.e., $\judgef_R(s, a, s')>0,  \forall{s\in S\backslash G, s'\in S, a\in A}$.
This is a standard assumption for SSPs~\cite{bertsekas1991analysis}.
There is a budget $\budget \in \mathbb{R}^+$ which we interpret as the cost an agent is willing to spend on morality before reaching a goal state $s \in \goals$.

One potential issue when integrating non-moral objectives and moral considerations is the presence of \emph{positive moral loops}.
Take the example of an autonomous system managing investments for a private business.
A policy ought to invest funds ethically while achieving goals, otherwise stakeholders will not adopt it.
A truly ethical policy may get caught in a positive moral loop and never achieve its goals (donating to charity until bankruptcy).
Thus, we prune policies over budget $\budget$.
Stationary policies $\pi:S\rightarrow A$, typical for SSPs, would prune moral loops entirely once they exceed the budget (no money for charity), which is why we chose non-stationary solutions $\policy: S \times \{0,\ldots,H-1\} \rightarrow A$.
Our solutions follow the moral loop for some number of cycles while leaving budget to pursue the goal afterwards.
The budget is similar to the bound vector used for Multi-Objective SSPs~\cite{chen2023heuristic}, though we bound solutions with respect to just one consideration instead of all objectives.

Solutions to infinite horizon SSPs are assumed to be proper, meaning the probability of reaching a goal from $s_0$ is 1.0.
In finite horizon problems, stochasticity  may prevent the policy from reaching a goal within the horizon. Thus, we say proper policies reach a goal from $s_0$ under the budget $b$ with non-zero probability.
\begin{assumption}[Weak Budgeted Improper Policies]

    For a MMSSP, all improper policies $\policy$ have an expected cost over the budget: $$\gatherf_R^{W^{\policy}_R[1]}(s_0,\policy(s_0,0))>b$$
    
    There exists a proper policy $\policy$ with an expected cost under the budget: $$\gatherf_R^{W^{\policy}_R[1]}(s_0,\policy(s_0,0))\leq b$$
\end{assumption}
We define \emph{optimal moral non-stationary policies} for a MMSSP:
\begin{align*}
  \Pi^b = \{\policy \in \Pi : \gatherf_R^{\pi}(s_0,\policy(s_0,0))\leq\budget\},\\
  \Pi^{b*} = \{\policy : \mathcal{N}(\policy)=\min_{\policy' \in \Pi^b}(\mathcal{N}(\policy')) \}, 
  \pi^{*} = \arg\min_{\policy \in \Pi^{b*}} \gatherf_R^{\pi}(s_0,\policy(s_0,0))
\end{align*}
The set $\Pi^{b}$ are all the \emph{proper policies} whose expected cost is under the budget $b$.
The set of \emph{moral proper policies} $\Pi^{b*}$ are those proper policies with minimal non-acceptability.
The \emph{optimal moral policy} is the moral proper policy with minimal expected cost.
\section{Moral Planner}
\label{sec:algorithm}
In the previous sections we construct a Multi-Moral MDP/SSP like a Multi-Objective MDP, with added lexicographic preferences and greater freedom over types.
We presented a mechanism to evaluate solution policies ethically using MEHR.
The following presents a heuristic moral planning algorithm with two stages.
First, an AO* Multi-Objective heuristic planning algorithm is adapted from \cite{chen2023heuristic} to find the set of proper \emph{Pareto undominated} policies.
Second, the undominated policies are evaluated using MEHR argumentation.
A policy with minimal non-acceptability (and then minimal expected cost) is output.
We give pseudocode in Algorithm~\ref{alg:MPlan}.
\begin{algorithm}[t]
    \algdef{SE}[DOWHILE]{Do}{doWhile}{\algorithmicdo}[1]{\algorithmicwhile\ #1}
\begin{minipage}[t]{\textwidth}
  \raggedright
  \textbf{Input}: Multi-Moral SSP $\langle S, A, P, s_0, H, M, C, L, R, \budget, \goals \rangle$, heuristic $\heuristic$\\
  \textbf{Output}: Optimal moral policy, \policy
\end{minipage}

\begin{algorithmic}[1]
\State$\vec{W} \gets \{ \{ \heuristic_c(s,t) : \forall{c\in C}\} \}$; $F \gets \{(s_0,0)\}$; $I \gets \emptyset;$
\State $\alpha(s,t) \gets \emptyset : \forall_{s\in S, t \in \{0,...,H-1\}}; \alpha(s_0, t) \gets A$
\Do
    \State $\vec{W}' \gets \vec{W}$; $Z \gets \text{postOrderDFS}(I\cup F, \alpha)$
    \For{$s, t \in Z$}
        \State$\vec{W}[t](s) \gets \{ \vgatherf^{\boldsymbol{W}[t+1]}(s,a) :\forall{a \in A, \boldsymbol{W} \text{ from } \vec{W}} \}$
        \State $\vec{W}[t](s) \gets \{\vec{w} \in \vec{W}[t](s) : \vec{w}_R[t](s) < \budget \}$
        \State $\vec{W}[t](s) \gets \text{pprune}(\vec{W}[t](s))$
        \State $\alpha[t](s) \gets \{a \in A : \vgatherf^{\boldsymbol{W}[t+1]}(s,a) \in \vec{W}[t](s) \}$
        \State $F \gets (F \backslash \{(s,t)\}); I \gets I \cup \{(s,t)\}$
        
        \State $F \gets F \cup \{(s',t+1) \backslash I : t+1<H \land \exists_{a\in A}P(s,a,s')>0 \}$
         
    \EndFor
\doWhile{$((F \cap Z) \backslash G = \emptyset) \land (\vec{W} \approx \vec{W'})$}
\State $\Pi \gets$ Extract policies $\pi$ from $\vec{W}$
\State $\phi^\pi \gets$ Extract Histories from $\pi$, $\forall{\pi \in \Pi}$
\State \Return $\policy \gets \text{MEHR}(\Pi, \phi)$
\end{algorithmic}

    \caption{MPlan Algorithm}
    \label{alg:MPlan}
\end{algorithm}
The algorithm explores the MMMDP as a graph whose nodes are state-time pairs.
It maintains a best partial sub-graph, represented by interior $I$, with successors inside the sub-graph, and a fringe $F$ with successors outside the sub-graph.
The best partial sub-graph expands/prunes state-time nodes if they potentially form an undominated solution.
A node's worth across moral considerations is grouped by a \emph{worth vector}, $\vec{w} \in 2^{\estimateType_1 \times \ldots \times \estimateType_{|C|}}$ where $\vec{w}_i$ is the expected worth for moral consideration $C_i$ (see Figure~\ref{fig:notation}).
A worth vector \emph{Pareto dominates} another if it is preferable by one moral consideration (including the cost function) and at least as preferable by all others. This ignores the weak lexicographic order.
\begin{definition}[Worth Vector Pareto Dominance]
    For two worth vectors $\vec{w}$ and $\vec{w}'$, then $\vec{w}$ Pareto dominates $\vec{w}'$ if and only if $\exists{i \in \{1,\ldots,|C|\}}, \vec{w}_i \succ_i \vec{w}'_i$ and $\forall{j \in \{1,...,|C|\}}, \vec{w}_j \succeq_j \vec{w}'_j$.
\end{definition}
Each state-time maps to a set of worth vectors, $\vec{W} : S \times \{0,...,H-1\} \rightarrow 2^{\{\estimateType_1 \times \ldots \times \estimateType_{|C|}\}}$ (see Figure~\ref{fig:notation}).
Each worth vector $\vec{w} \in \vec{W}[t](s)$ represents the expectations of a potential undominated solution at time $t$, state $s$.
On line 1, $\vec{W}$ gets one worth vector for each node from a heuristic estimate of the consideration optimal worth function $W^*_c[t](s)$ (Section \ref{sec:MMMDP}), $\heuristic_c : S \times [0,...,H-1] \rightarrow \estimateType_c$.
Initially, frontier $F$ has initial state $s_0$ at time $0$ and interior $I$ is empty (line 1).
The sub-graph will grow to contain state-time nodes reachable in undominated solutions. Specifically, $\alpha: S \times \{0,\ldots,H-1\} \rightarrow 2^A$ maps a state-time to their mapped actions from all potential undominated solutions, Initially, any solution could be undominated, so $\alpha$ maps to all actions for state $s_0$, time $0$.

The algorithm performs iterative \emph{back-ups} on state-time nodes in the sub-graph. This informs the action mappings of potential solutions $\alpha$.
The sub-graph is backed up via post-order DFS or descending time order (line 4-5) to maximise each back-up's effect.

A back-up on a state-time collects the multi-worth function $\boldsymbol{W}$ (Section~\ref{sec:MMMDP}) from each permutation of expected worth in $\vec{W}$, expressed in pseudocode as `$\forall{\boldsymbol{W} \text{ from } \vec{W}}$' (line 6).
In an informal example, say $\vec{W}[t](s)$ contains two worth vectors, estimating the worth of two potential undominated solutions for state $s$, time $t$, $\vec{W}[t](s) = \{\vec{w}_1, \vec{w}_2\}$.
Given one successor with $\vec{W}[t+1](s') = \{\vec{w}'_1, \vec{w}'_2\}$, 4 multi-worth functions are constructed
($\vec{w}_1+\vec{w}'_1$, $\vec{w}_1+\vec{w}'_2$, $\vec{w}_1+\vec{w}'_2$, $\vec{w}_2+\vec{w}'_2$), placing estimate permutations for state $s$, time $t$ and state $s'$, time $t+1$ into $\boldsymbol{W}$.
In a back-up, these multi-worth functions are passed to the worth aggregation function for every action. This finds the worth vector expectation of applying each action to a potential undominated solution. For MMSSPs, worth vectors over the budget are pruned away (line 7). The \textit{pprune} algorithm~\cite{roijers2017multi_pprune} prunes all Pareto dominated worth vectors (line 8).

The actions that generated undominated worth vectors are added to $\alpha[t](s)$ (line 9).
After a back-up, fringe nodes are expanded, moved from the fringe to the interior and replaced by successors (lines 10-11).
The algorithm reiterates over the expanded sub-graph (line 3), backing-up state-time nodes reachable through actions in $\alpha$, thus, nodes that are in-budget and promising by some moral consideration. Other nodes are pruned.
This focused search avoids full sweeps across the state space, as required by planning algorithms like (MO) Value Iteration~\cite{Bellman_DynamicProgramming_57,white1982multi}.

Iteration continues until there are no unexpanded state-time nodes that could be in an undominated solution (line 12).
Additionally, the expectations of solutions must be consistent between iterations.
To find this, we generalise the $\approx$ operator from the moral considerations to $\vec{W}$ and a copy before each iteration, $\vec{W}'$.
We say $\vec{W} \approx \vec{W}'$ if and only if $\forall{s\in S, t\in\{0,\ldots,H\}}, \exists \vec{w} \in \vec{W}$ and $\vec{w'} \in \vec{W'}$ such that $\forall{c\in C}, \vec{w}_c \approx \vec{w'}_c$.
In other words, each worth vector from $\vec{W}$ must correspond to an equivalent in $\vec{W}'$.

After a set of Pareto undominated worth vectors are found, all policies are extracted on line 13 (algorithm omitted for space).
The worth and probability of policy histories is found on line 16 using Algorithm~\ref{alg:getOutcomeWorths}.
These values are passed to the MEHR argumentation procedure to find the most preferable policies.
This is implemented with the algorithm described in~\cite{kolker2023uncertain}, with actions as policies, potential branches of future development as histories, and moral theories as described in Section~\ref{sec:MEHR}.

\section{Experiment}
We evaluate our algorithm with experiments based on The Lost Insulin example.
Implementation, MDPs and results are available on \url{https://github.com/sameysimon/MoralPlanner}.
All experiments ran on a 2020, M1 Macbook Air with 8GB of memory.

We expand the scenario for the experiments. Hal urgently needs insulin. There is a scheduled resupply in $200$ minutes. Each time step represents a $10$ minute increment. At each step, there is a $0.6$ probability Hal dies without insulin.
Hal can wait or go to Carla's house. There is a $0.2$ chance he is caught, arrested and forced to wait.
If Hal gets to Carla's, he will begin searching, though he can go home and wait.
While searching, he may compensate Carla: he can leave a lot of money with a $0.7$ chance of covering the insulin cost; he can leave a small amount, with a $0.1$ chance of covering the cost.
Once finding the insulin, he can steal it or leave and go home.
Without insulin, Carla has $0.1$ chance of dying at each time step.
We assume Hal replaces Carla's insulin after the delivery.

We discern a few moral theories. As before, a collective Act-Utilitarianism measures Hal and Carla's well-being. This can divide into \textit{Ethical Egoism}, measuring only Hal's utility~\cite{sep-egoism}, and \textit{Altruism}~\cite{sep-altruism}, measuring only Carla's wellbeing.
There is an absolute theory against stealing and another that allows stealing if there is compensation.
For MMSSPs, states where Hal has insulin are goals and all transitions cost 1. We set the budget $b$ to $18.5$.

We use domain dependent heuristics for $\heuristic(s,t)$. We estimate Hal's utility with $-0.4$ when he waits and $0$ otherwise; Carla's utility to $-0.1$ when she loses her insulin and $0$ otherwise. For absolute theories, we estimate $\bot$ since Hal can always avoid stealing.

The resulting MDP has 286 state-time pairs. We ran configurations with different moral theories and lexicographic ranks 5 times each and averaged the total run time. Results are in Table~\ref{tab:MoralPlanner}.
\begin{table}[t]
    \centering
    \scalebox{1}{
\begin{tabular}{|l|llllll|}
\hline
\multicolumn{1}{|c|}{\multirow{3}{*}{Config.}} & \multicolumn{6}{c|}{Moral Planner}                                                                                                                                                                                                                                                                                                                                                \\ \cline{2-7} 
\multicolumn{1}{|c|}{}                         & \multicolumn{3}{c|}{$\gatherf^{W[1]}(s_0, \pi(s_0,0))$}                                   & \multicolumn{1}{c|}{\multirow{2}{*}{\begin{tabular}[c]{@{}c@{}}Non-\\ acc.\end{tabular}}} & \multicolumn{1}{c|}{\multirow{2}{*}{\begin{tabular}[c]{@{}c@{}}No. of\\ solns.\end{tabular}}} & \multicolumn{1}{c|}{\multirow{2}{*}{\begin{tabular}[c]{@{}c@{}}Time\\ (ms)\end{tabular}}} \\ \cline{2-4}
\multicolumn{1}{|c|}{}                         & \multicolumn{1}{c|}{Hal}     & \multicolumn{1}{c|}{Carla}   & \multicolumn{1}{c|}{Steal}  & \multicolumn{1}{c|}{}                                                                     & \multicolumn{1}{c|}{}                                                                         & \multicolumn{1}{c|}{}                                                                     \\ \midrule
$C^0, H^0$                                     & \multicolumn{1}{l|}{-8.9312} & \multicolumn{1}{l|}{-1.1007} & \multicolumn{1}{l|}{N/A}    & \multicolumn{1}{l|}{0.109}                                                                & \multicolumn{1}{l|}{8}                                                                        & 52.2                                                                                      \\ \hline
$C^0, H^1$                                     & \multicolumn{1}{l|}{-10.211} & \multicolumn{1}{l|}{0}       & \multicolumn{1}{l|}{N/A}    & \multicolumn{1}{l|}{0}                                                                    & \multicolumn{1}{l|}{8}                                                                        & 53.0                                                                                      \\ \hline
$C^1, H^0$                                     & \multicolumn{1}{l|}{-8.9312} & \multicolumn{1}{l|}{-1.1007} & \multicolumn{1}{l|}{N/A}    & \multicolumn{1}{l|}{0}                                                                    & \multicolumn{1}{l|}{8}                                                                       & 66.4                                                                                      \\ \hline
$C^0, H^0, S^0$                                & \multicolumn{1}{l|}{-8.9312} & \multicolumn{1}{l|}{-1.1007} & \multicolumn{1}{l|}{$\top$} & \multicolumn{1}{l|}{0.237}                                                                & \multicolumn{1}{l|}{8}                                                                        & 70.2                                                                                      \\ \hline
$C^1, H^0, SC^0$                               & \multicolumn{1}{l|}{-8.9312} & \multicolumn{1}{l|}{-1.1007} & \multicolumn{1}{l|}{$\top$} & \multicolumn{1}{l|}{0.147}                                                                & \multicolumn{1}{l|}{8}                                                                       & 77.0                                                                                      \\ \hline
$C^0, R$                                       & \multicolumn{1}{l|}{-18.387} & \multicolumn{1}{l|}{-0.7705} & \multicolumn{1}{l|}{N/A}    & \multicolumn{1}{l|}{0}                                                                    & \multicolumn{1}{l|}{4}                                                                        & 52.1                                                                                      \\ \hline
$C^0, S^0, R$                                  & \multicolumn{1}{l|}{-18.387} & \multicolumn{1}{l|}{-0.7705} & \multicolumn{1}{l|}{$\top$} & \multicolumn{1}{l|}{0}                                                                    & \multicolumn{1}{l|}{4}                                                                        & 62.5                                                                                      \\ \hline
\end{tabular}

    }
    \caption{Moral Theory for Hal's Ethical Egoism denoted by $H$; Carla's Altruism by $C$; absolute no stealing by $S$; stealing allowed with compensation by $SC$; non-moral cost function by $R$. The superscript denotes lexicographic rank. Shows config non-acceptability and execution time to 3 s.f.}
    \label{tab:MoralPlanner}
\end{table}
When Hal and Carla's theories have equal rank, a policy for stealing the insulin is selected, matching intuition since Hal's life is at greater risk. 
The non-acceptability is greater than 0 meaning there is a moral dilemma. This is likely because it is not guaranteed that Hal's life will be saved. From the perspective of the outcomes where Hal steals the insulin but dies anyway, there would be negative retrospection.
High and low compensation policies yield the same non-acceptability as these theories are ambivalent.
When Hal's life is lexicographic preferred, any chance of saving Hal is worth risking Carla.
When Carla's life is lexicographic preferred, Hal must wait at home since any risk to Carla is not permissible. 

Introducing the law against stealing, the policy where Hal steals insulin is still preferable though non-acceptability has risen. This is because more outcomes are indefensible.
With the compensation theory, Hal leaves the maximum amount and non-acceptability somewhat recovers.
Considering the MMSSP examples, in both cases, a middle ground that is best for Carla while remaining in budget is selected.
We see a trend across the results that a greater number of moral considerations leads to a greater running time.

In Table 2, we show total back-ups, iterations and state expansions.
Absolute moral theories back propagate easily due the simplicity of the moral consideration.
The non-moral cost function causes more backups, but reduces the number of iterations. This may be because search is directed towards policies that go to Carla's house, which contains more states, but makes convergence faster.
\begin{table}
    \centering
    \begin{tabular}{|c|c|c|c|}
    \hline Config. & Expanded & Backups & Iterations \\ \midrule
    Hal \& Carla & 243 (84\%) & 2472 & 23 \\ \hline
    Hal \& Carla \& Steal & 243 (84\%) & 2472 & 23 \\ \hline
    Hal \& Carla \& Steal\_Comp & 243 (84\%) & 2472 & 23 \\ \hline
    Non-Moral Hal \& Carla & 271 (95\%) & 2822 & 21 \\ \hline        
\end{tabular}
    \caption{Table showing number of state-time expansions (with \% of 286 state-time space), backups and iterations performed by the Multi-Moral AO* algorithm.}
    \label{tab:my_label}
\end{table}

\section{Discussion}
We have formalised moral theories for ethical planning with uncertainty over decision outcomes and moral theories.
While the theories in this paper are simple (expected utility maximisation and probabilistic absolutism), we hope our formalism is expressive.
One theory could back-propagate all state-probability information in its worth type $\estimateType$, instead of reducing to a `single expected' worth. This represents a great deal at the cost of exponential blowup. In future, we will explore the balance of tractable expression.

Our approach is compatible with non-moral goals and costs with MMSSPs.
We showed solutions must be non-stationary to account for positive moral loops. Future work may explore eliminating these loops (like traps~\cite{kolobov2011heuristic}) to alleviate this.
Our algorithm finds a proper policy if one exists, unlike
the work by~\citeauthor{svegliato2021ethically} which modifies ethical polices towards non-moral goals. If this can not be done under a threshold, no policy is returned~\cite{svegliato2021ethically}. In future, we could try to apply this approach to our algorithm.

In some MMMDPs, there is a large number of undominated policies. Some multi-objective algorithms find the Pareto Front or the Convex Coverage Set (CCS) which is generally smaller~\cite{roijers2017multi}.
Unfortunately, these may not include the MEHR optimal policy.
Additionally, policies can have an exponential number of histories and since MEHR compares each policy's arguments, we expect time+space complexity to be exponential. 
Developing an n-step lookahead algorithm would help and 
we may try Monte Carlo history sampling to limit the number of arguments. 
We plan to search for optimisations and recommend our current algorithm for small scale problems which require more ethical scrutiny.

We hope our novel formalism helps the bridge between probabilistic planning and the expressive realm of Ethics. We integrated ethical heuristics to guide our search, which we believe is novel for \textit{Machine Ethics}.
The approach can express moral uncertainty, shown by experiments sensitive to configurations of moral theories.
Our approach has potential for explainability with MEHR decisions based on arguments that are reproducible following an inquiry.
\section*{Acknowledgements}
We would like to thank the University of Manchester for funding and EPSRC under the Centre for Robotic Autonomy and Demanding and Long Lasting Environments (CRADLE) (EP//X02489X/1).
\bibliographystyle{ieee} 
\bibliography{refs}

\end{document}